\documentclass[runningheads]{llncs}

\usepackage{graphicx}
\usepackage{tabularx}
\usepackage{subfig}
\usepackage{arydshln}
\usepackage{soul, xcolor}
\usepackage{multirow}
\usepackage{todonotes}
\usepackage{hyperref}
\usepackage[misc,geometry]{ifsym}

\newcolumntype{H}{>{\setbox0=\hbox\bgroup}c<{\egroup}@{}}

\definecolor{LightCyan}{rgb}{0.88,1,1}

\newcommand{\model}{\texttt{DetONADe}}
\newcommand{\dataset}{\texttt{ODIN}}

\newcolumntype{P}[1]{>{\centering\arraybackslash}p{#1}}
\newcolumntype{M}[1]{>{\centering\arraybackslash}m{#1}}
\begin{document}

\newcommand{\repeatthanks}{\textsuperscript{\thefootnote}}

\title{Detecting Anchors' Opinion \\in Hinglish News Delivery}

 \author{Siddharth Sadhwani\thanks{Equal Contribution}$^{\textrm{\Letter}}$ \and Nishant Grover\repeatthanks \and Md Shad Akhtar \and Tanmoy Chakraborty} 

 \institute{Dept. of CSE, IIIT-Delhi, India \\ \email{\{siddharth18313, nishant18399, shad.akhtar, tanmoy\}@iiitd.ac.in}}

\maketitle

\begin{abstract}
Humans like to express their opinions and crave the opinions of others.  Mining and detecting opinions from various sources are beneficial to individuals, organisations, and even governments.  One such organisation is news media, where a general norm is not to showcase opinions from their side. Anchors are the face of the digital media, and it is required for them not to be opinionated. However, at times, they diverge from the accepted norm and insert their opinions into otherwise straightforward news reports, either purposefully or unintentionally. This is primarily seen in debates as it requires the anchors to be spontaneous, thus making them vulnerable to add their opinions. The consequence of such mishappening might lead to biased news or even supporting a certain agenda at the worst. To this end, we propose a novel task of anchors' opinion detection in debates. We curate code-mixed news debates and develop the \dataset\ dataset. A total of 2054 anchors' utterances in the dataset are marked as opinionated or non-opinionated. Lastly, we propose \model\ -- an interactive attention-based framework for classifying anchors’ utterances and obtain the best weighted-F1 score of 0.703. A thorough analysis and evaluation show many interesting patterns in the dataset and predictions.

\keywords{Anchors' opinion\and Opinion Detection \and Code-Mixed Conversations.}
\end{abstract}
\section{Introduction}

News bulletins play a  significant role in educating, informing, spreading awareness, and influencing the masses about important current affairs. Recent estimates show that the Indian news channels are broadcast over 161 million TV households, and more than 200 million internet users \cite{kpmg}. This puts a lot of responsibility on the news channels as they are the primary source of the masses' knowledge about current affairs. 

Common citizens expect their news to be free of opinions and only based on facts. However, in recent years, we have seen countless instances where reporters/news anchors either purposefully or unintentionally insert their opinions into otherwise straightforward news articles, thus earning the tags of biased news or biased reporting. They do so by blurring the line between fact-based news reporting and edited-agenda based news reporting. As a consequence, readers and viewers often get confused or get exposed to the targeted viewpoints of others. This practice is even more prevalent in live news reporting or during news debates wherein the content on display is unstructured and somewhat spontaneous compared to written news articles that get processed by many people, including editors, content managers, etc., making the reporter vulnerable to voice out their opinions. To this end, we propose {\bf a novel task of anchor's opinion detection in code-mixed conversations}. The objective of the task is to identify the anchor's utterance when they posit their own opinion in between. Table~\ref{tab:dataset:example} shows one such instance where the anchor appends their opinion in a news debate. The highlighted blue text signifies the opinionated spans. 

\begin{table*}[t]
    \centering
    \caption{An annotated snippet of a dialog (debate) in \dataset. For brevity, we do not show the full conversation. Anchor's opinion spans are highlighted in blue. [A]: Anchor's utterance, whereas, [$\mathbf{S_j}$]: \textit{j}th speaker's utterance.}
    \label{tab:dataset:example}
    \renewcommand{\arraystretch}{1.24}
    \resizebox{\textwidth}{!}{%
    \begin{tabular}{Hp{40em}|c}

        & \textbf{Utterance} & \textbf{Opinion}\\ \hline \hline 
        & [$\mathbf{S_1}$] $\cdots$ & \\ \hdashline
        \multirow{7}[40]{*}{Hinglish} & [\textbf{A}] \textit{``Nahi aap aap party ke prvakta ke taur par baithe hai ya political analyst vkhyatigat roop se baithe hai?''} (Are you representing yourself as a party representative or a political analyst?) & No \\ \hdashline
        & [$\mathbf{S_2}$] \textit{``maine aapse kya kaha sabko is laxman rekha ka samman karna chahiye yeh nirdesh sab par prabhavi roop se lagu hoga''} (What did I say to you? We all have to respect the rule, this rule will be implemented effectively) & - \\ \hdashline
        & \textbf{[A]} \textit{``\textcolor{blue}{aap thodi na teh karenge jo aapko apmaanjanak lag jaye kuch maine puch liya meri nazar main nahi hai} \underline{$<$name$>$} \underline{$<$name$>$} \underline{$<$name$>$} \underline{$<$name$>$} \underline{$<$name$>$} \underline{$<$name$>$} aap abhi lage huye hai brashtachaar ke bhism pitamah doordanth apradhiyo ke sanghrakshan karta $<$name$>$ avedh sarkar ke kamjoor mukhiya tanashah \textcolor{blue}{aap log abhi bhi lage huye hai aap log sudhar hi nahi rahe hai $<$name$>$ itna samjha rahe aapko}''} (You won't decide even if you feel that something is disrespectful, if it is not in my eyes $<$name$>$ sir $<$name$>$ sir $<$name$>$ sir $<$name$>$ sir $<$name$>$ sir $<$name$>$ you are also involved in corruption. You are still not understanding, $<$name$>$ is trying so hard to make you understand) & Yes \\ \hdashline
        & [$\mathbf{S_3}$] $\cdots$ & - \\ \hdashline
        & \textbf{[A]} \textit{``\textcolor{blue}{to ab jail bhejiyega na aapko bhi fayda milega unko jail bhejiyega ab agar woh aisa kuch kahe jail bhejiyega unko}''} (Send him to jail you will also profit from it. Send him to jail if he says something like this again.) & Yes \\ \hdashline
        & [$\mathbf{S_3}$] \textit{``pehle sun lijiye to''} (Listen to me first) & - \\ \hdashline
        & \textbf{[A]} \textit{``\textcolor{blue}{sab ek doosre ko jail bhejiyega sab mile huye hai aapne koi kasar thodi na chhodi hai aap logo ne kya kuch kaha hai $<$name$>$ ji yeh to ab aapko bhi fayda milega unko jail bhejiyega}''} (Send each other to jail. You all are in a cohoots. You have'nt left any stone unturned, you people have said too much, $<$name$>$ sir, now you will also get the benefit, send them to jail) & Yes \\ \hdashline
        & [$\mathbf{S_1}$] $\cdots$ & \\ 
        \hline
        
    \end{tabular}%
    }
    \vspace{-6mm}
\end{table*}

Such utterances should not be presented by an anchor, as the reporter is the face of the news media, and news media is supposed to be unbiased and unopinionated. To help and cater to the news anchors and media in promoting unbiased and unopinionated news, our work aims to help organisations detect opinionated utterances. To this end, we first curate \dataset\ (\textbf{O}pinion \textbf{D}etection \textbf{I}n \textbf{N}ews) -- a dataset developed by transcribing different code-mixed Hinglish news debates from several mainstream national news channels and annotating the utterances of the debate as opinionated/unopinionated. 
Then, we propose \model\ (\textbf{Det}ecting \textbf{O}pinion in \textbf{N}ews \textbf{A}nchor \textbf{De}livery) to detect utterances as opinionated and benchmark the task. We also present a detailed analysis of the dataset and necessary evaluation of the obtained results.

In summary, we make the following contributions:
\begin{enumerate}
  \item We explore opinion detection in a code-mixed (Hinglish) dialogue environment, a \textbf{novel task} which, to the best of our knowledge, has never been attempted before.
  \item We curate a \textbf{new dataset}, \dataset, by transcribing various Hinglish news debates from three national news channels and annotating these captioned utterances as opinionated/unopinionated.
  \item We perform \textbf{extensive analysis} of \dataset\ and provide interesting insights.
  \item We benchmark \dataset\ using \model\ and report the necessary results and error analyses.
\end{enumerate}
The source codes and datasets are available at \url{https://github.com/LCS2-IIITD/ODIN-PAKDD}.

\section{Related Work}

Opinion expression is an integral part of opinion mining, and it was first defined as either Direct Subjective Expression (DSEs) or Expressive Subjective Expressions (ESEs) \cite{wiebe_annotating_2005}. Following this definition, a fine-grained opinion Mining corpus, namely Multi-Perspective Question Answering (MPQA) was curated for annotating expressions as opinion. Apart from already present datasets, researchers also explored social media, blogs and news articles as opinion mining from heterogeneous information sources can be of great use for individuals, organisations or governments. Ku et al. \cite{lun-wei} dealt with the task of opinion extraction, summarisation and tracking on news and blogs corpora, and a lexicon-based feature modelling technique was proposed to extract opinions from documents. Support Vector Machines (SVM) and Decision Trees (C5) were used to predict the results. Breck et al. \cite{breck2007identifying} proposed a  Conditional Random Fields (CRF) based model where identifying opinion expression was assumed as a sequence labelling task and achieved expression-level performance within 5\% of human inter-annotator agreement. Raina \cite{raina} proposed an opinion mining model that leveraged common-sense knowledge from ConceptNet and SenticNet to perform sentiment analysis in news articles, achieving an F1-score of 59\% and 66\% for positive and negative sentences, respectively. Recently, researchers explored deep learning for opinion detection \cite{irsoy-cardie-2014-opinion,xie}.

The scope of this task has always been limited to English language and monolingual settings. There has not been any significant research work on opinion detection in code-mix or Indic languages. But code-mixing is an increasingly common occurrence in today's multilingual society and poses a considerable challenge in various NLP based downstream tasks. Accordingly, there have been some helpful developments in the field of code-mix in the form of sentiment analysis task, humour, sarcasm and hate-speech detection. A dual encoder based model for Sentiment Analysis on code-mixed data was proposed wherein the network consisted of two parallel BiLSTMs, namely the collective and the specific encoder \cite{lal}. This model particularly generated sub-word level embeddings with the help of  Convolutional Neural Networks (CNNs) to capture the grammar of code-mixed words. Recently, pretrained monolingual and cross-lingual deep learning models were also leveraged for detection of hate-speech and sarcasm on code-mixed data \cite{pant} wherein they used fine-tuned RoBERTa and ULMFit for English and Hindi data streams, respectively. For cross-lingual setting, XLM-RoBERTa was fine-tuned on transliterated Hindi to Devanagri text.

The works mentioned above do leverage code-mix text for common downstream tasks. However, no research has been done on opinion detection on code-mix text in sequential data streams. Most opinion detection and sentiment analysis studies have focused on news articles, blogs and movie reviews. In online news articles, every piece is reviewed by multiple people, and thus the scope of opinions is limited compared to news coverage on live media sources. Moreover, biased and opinionated live news anchoring can significantly impact our society and go against the essence of free and fair news reporting. Therefore, we aim to detect opinions amongst news anchors. We focus on code-mix news anchoring mainly in live video debates through national news channels that stick to a mix of Hindi and English language (Hinglish) for news distribution. Moreover, our deep learning model not only leverages context but does so in a sequential manner, thus focusing on the text and the utterances before a statement to classify the text as accurately and robustly as possible.

\section{Dataset}
\begin{figure}[t]
    \centering
    \includegraphics[width=\textwidth]{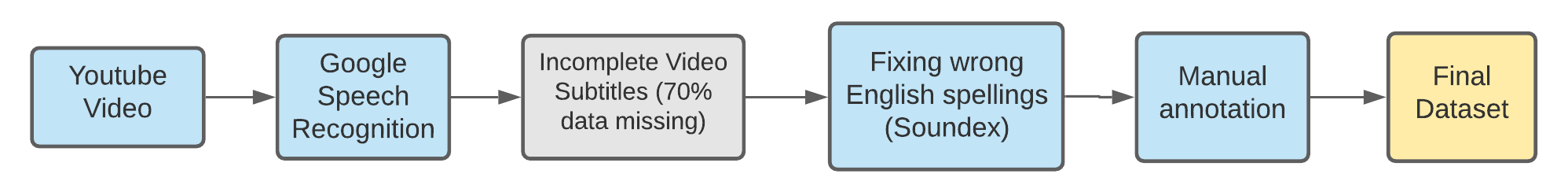}
    \vspace{-4mm}
    \caption{Dataset annotation pipeline.}
    \label{fig:pipe}
    \vspace{-4mm}
\end{figure}

In this section, we lay out the details of the dataset development process. First, we extract debate videos from three popular Indian Hindi news Youtube's channels -- ABP News\footnote{https://www.youtube.com/c/abpnews}, Aaj Tak\footnote{https://www.youtube.com/c/aajtak}, and Zee News\footnote{https://www.youtube.com/c/zeenews}. Subsequently, the collected videos were processed to extract the romanized Hinglish code-mixed utterances. Each utterance is uttered either by the anchor or by the invited speakers. To ensure sanity, we do not identify the utterances with the speaker's name; instead, we assign ids ($A$ for the anchor, and $\{S_1, S_2, \cdots, S_n\}$ for the invited speakers) to each utterance. Next, we annotate the anchor's utterances as opinionated\footnote{A personal sentiment, which describes the anchor's feeling on the topic \cite{affect-sentiment}.} or non-opinionated depending upon the dialog conversation. A high-level dataset development process is outlined in Figure \ref{fig:pipe}.

\paragraph{\bf Prepossessing.} We collect 46 debate\footnote{Henceforth, we will use debate, dialogue, and conversation interchangeably to signify a sequence of utterances.} videos for two broad topics as religious and political. Initially, we obtain transcriptions of these videos using the Google Speech Recognition\footnote{https://pypi.org/project/SpeechRecognition/} tool. The obtained output had many missing words, possibly due to the background noise or due to the code-mixed nature of the conversation; therefore, we manually add the missing words to complete the utterances. Furthermore, we observe many spelling mistakes in English words -- `\textit{laiv}' for `\textit{live}', `\textit{ophis}' for `\textit{office}', `\textit{daunalod}' for `\textit{download}', etc. To fix these spelling mistakes, we try mapping words in the English dictionary to the words in question based on word similarity. 
We use phonological similarity to achieve this. We employ Libindic’s Soundex library\footnote{https://libindic.org/Soundex} to obtain the correct mapping.

\begin{table}[t]
    \centering
    \caption{Dataset statistics of \dataset.}
    \resizebox{0.55\textwidth}{!}{
    \label{tab:dataset:stat}
    \begin{tabular}{l|c}
        \textbf{Features} & \textbf{Value} \\ \hline \hline
        Number of dialog (debate) videos & 46  \\ \hline
        Average length of the videos &  33 mins \\ \hline
        Number of utterances & 4490 \\ \hline
        Number of anchor utterances  & 2054  \\ \hline
        \bf Number of opinionated anchor utterances & \bf 597 \\ \hline
        Number of tokens &  261811 \\ \hline
        Number of unique tokens (vocabulary) & 20023 \\ \hline
        Average number of utterances per dialog &  97.6 \\ \hline
        Maximum number of utterances in a dialog & 233 \\ \hline
        Average number of words per utterance &  58 \\ \hline
        Maximum number of words in an utterance & 1192 \\ \hline
        
    \end{tabular}}
    \vspace{-3mm}
\end{table} 

\paragraph{\bf Annotation.} Each debate has a series of utterances -- some of them were uttered by the anchor and others by the invited speakers. We employ two annotators to annotate the anchor's utterances as opinionated or non-opinionated. Since the objective of the current work is to identify opinions of anchors, we do not annotate speaker's utterances. Both annotators read the utterances of the debate and annotate the whole data. To check the inter-rater agreement, we compute Cohen's $\kappa$ value of $0.88$. Subsequently, we perform a consolidated step to include only those annotations where both annotators agree on the opinionated label -- we treat disagreement as non-opinionated. Table \ref{tab:dataset:example} shows an annotated dialog. For brevity, we show only a snippet of the dialog. There are three speakers and one anchor debating over a topic. Out of all utterances in the dialog, we show the annotated anchor's utterances as opinionated and non-opinionated.

\paragraph{\bf Statistics.} A detailed statistic of \dataset\ is listed in Table \ref{tab:dataset:stat}. There are total 46 debate videos with an average length of $\sim$33 mins. In total, there are 4490 utterances -- 2054 anchor utterances and 2436 other speakers utterances. Out of 2054 anchor utterances, 597 of them are opinionated, accounting for approximately $30\%$ of the utterances. 

\begin{figure}[t]
    \centering
    \includegraphics[width=7.5cm]{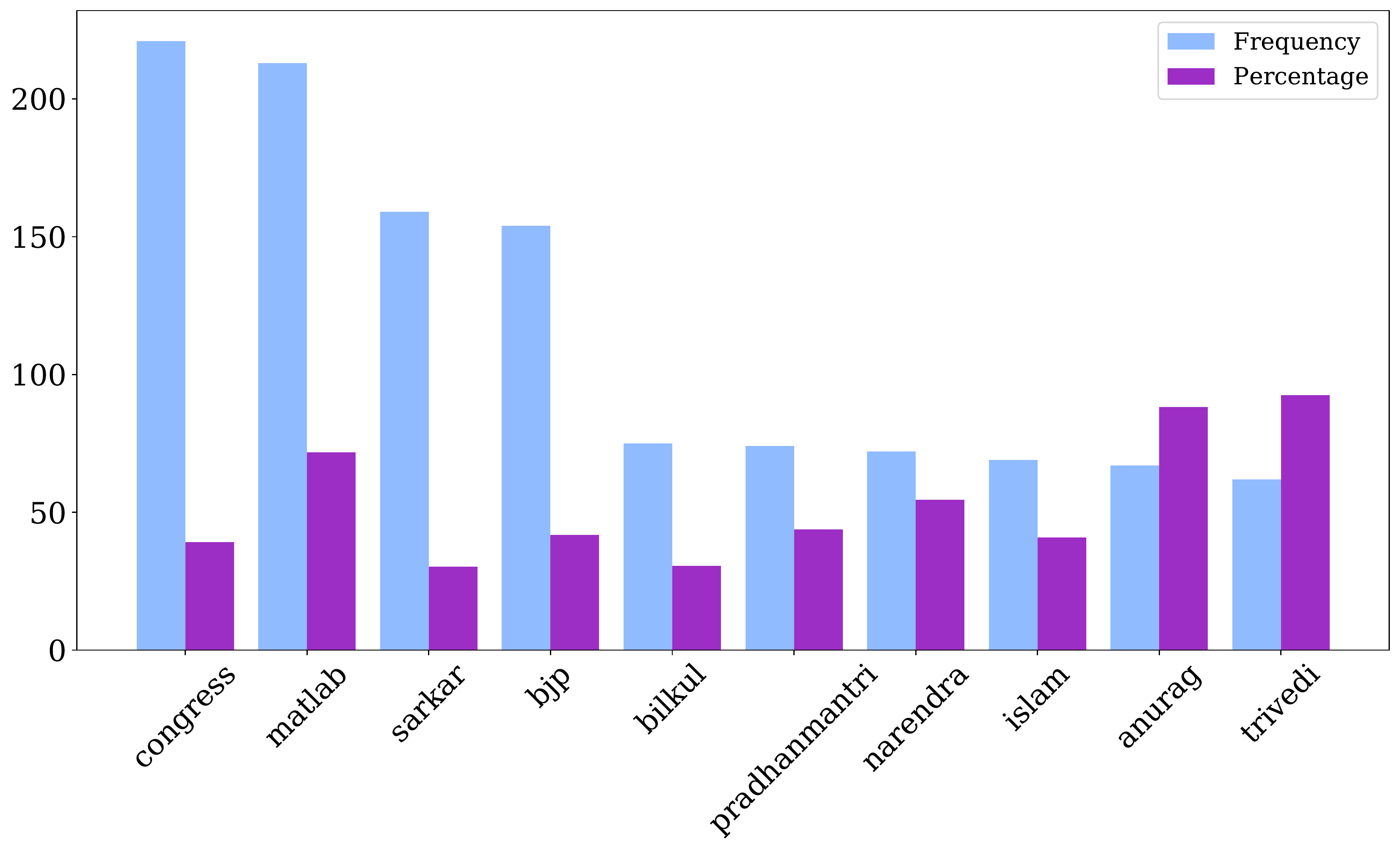}
    \vspace{-4mm}
    \caption{Top 10 most occurring words in opinionated anchor utterances.}
    \label{fig:top10OP}
    \vspace{-2mm}
\end{figure}

\begin{figure}[t]
\centering
\subfloat[Politics]{\includegraphics[width=0.48\linewidth]{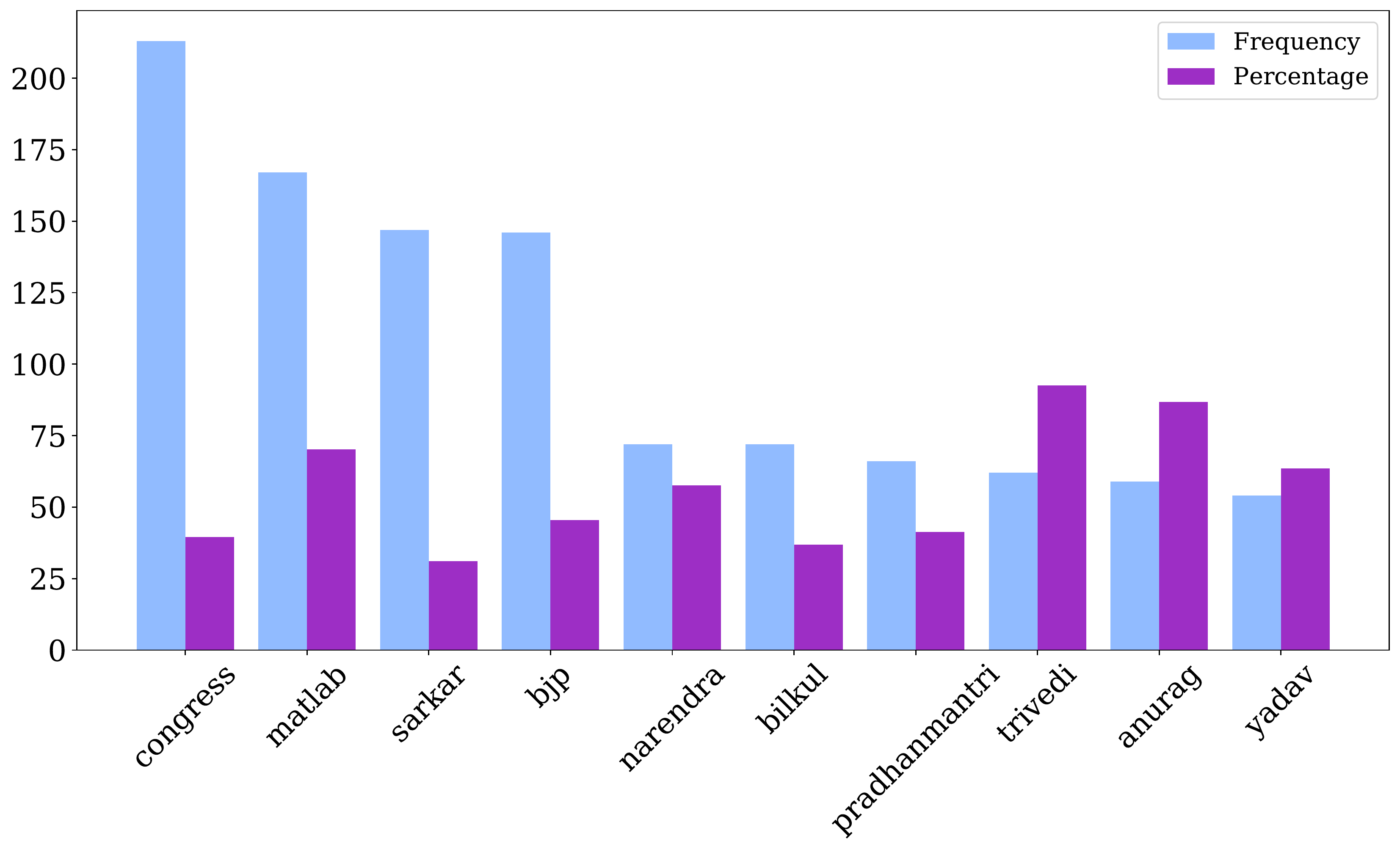}\label{fig:top10OPpolitics}}\quad
\subfloat[Religion]{\raisebox{1.6ex}{\includegraphics[width=0.48\linewidth]{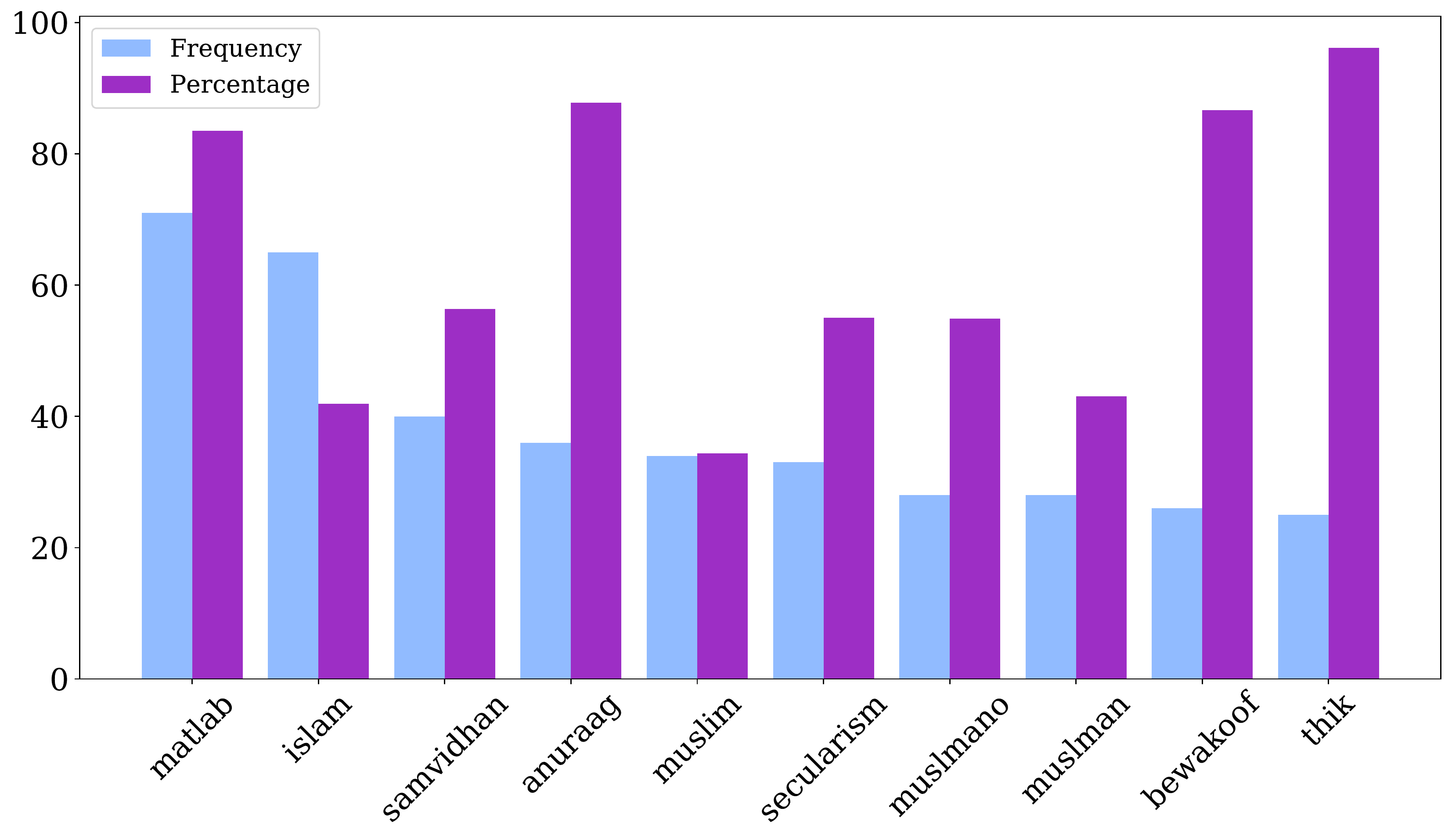}}\label{fig:top10OPreligion}}\\

\vspace{-3mm}
\caption{Topic-wise top 10 most occurring words in opinionated anchor utterances.}%
\vspace{-3mm}
\end{figure}

\paragraph{\bf Dataset Insights.} We analyze the dataset to gain insight of the inherent pattern in opinionated utterances. Apart from various topic-related terms such as \textit{BJP}\footnote{BJP and Congress are two major political parties in India.\label{lbl:note}}, \textit{Congress}\footnotemark{\ref{lbl:note}}, \textit{Islam}, etc., corresponding to the political and religious topics, we observe opinionated words like `\textit{bilkul}' (certainly), `\textit{matlab}' (means), `\textit{theek}' (ok), etc., have a significant presence in opinionated utterances. We depict a bar graph of the top-10 most frequent words in opinionated utterances in Figure \ref{fig:top10OP}. Moreover, to comprehend whether the frequent words are opinion specific or not, we also plot the ratio to see the distribution of these words in opinionated v/s all the utterances. In Figures \ref{fig:top10OPpolitics} and \ref{fig:top10OPreligion}, we observe the most frequent words in a topic-wise segregated form. 

We observe cases where two or more speakers are involved in a heated exchange without a concrete outcome. In such scenarios, the anchor tries to calm them down, and while doing so, the anchor often slide their own opinions on the subject matter. Cases like these involve the anchor repeatedly calling the name of the speakers -- we observe that $\sim33\%$ of the opinionated utterances have a single word (or name) spoken multiple times. One such example is shown in the anchor's second utterance in Table \ref{tab:dataset:example}. We also find out that anchors tend to ask more questions in an opinionated utterance -- on average, 1.45 question words are present in an opinionated utterance, whereas, only 0.9 question words are there in a non-opinionated utterance. 

\begin{figure}[t]
    \centering
    \includegraphics[width=7cm]{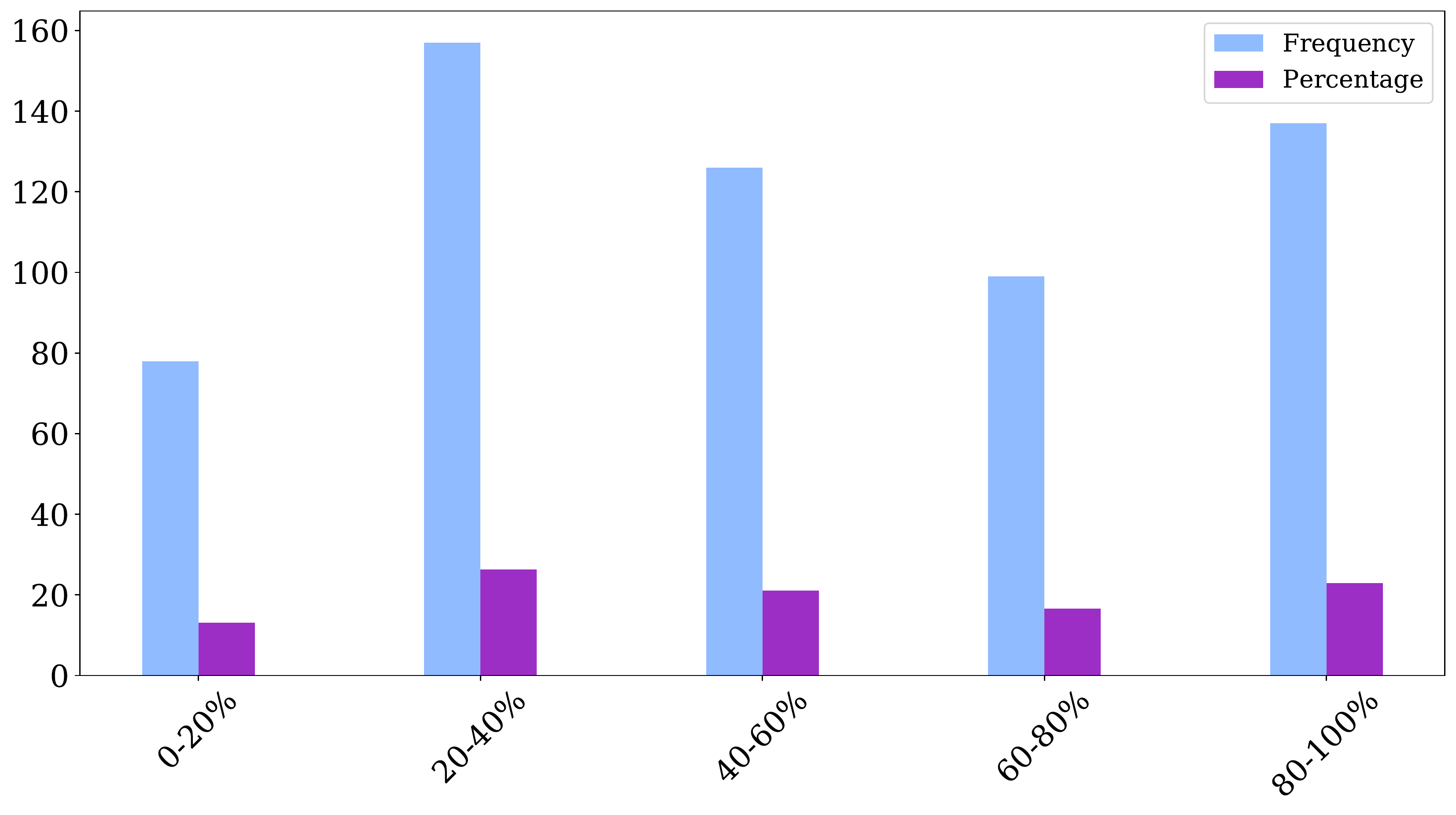}
    \vspace{-4mm}
    \caption{Time-wise distribution of anchor opinionated utterances. This signifies that an anchor is more conscious about expressing their opinions at the start of the debate and as the debate goes by they get more spontaneous and less conscious.}
    \label{fig:timewise}
    \vspace{-6mm}
\end{figure}

On careful analysis, we observe that an anchor is relatively more likely to express personal opinion at the later stage of the debate rather than at the beginning of the debate. We plot the distribution of the opinionated utterances on the time scale in Figure \ref{fig:timewise}. As we can see that only 78 utterances are opinionated during the first 20\% of the debate duration, which increases to 157 during 20-40\%, 126 during 40-60\%, 99 during 60-80\%, and 137 during 80-100\% of the debate duration.This signifies that an anchor is more conscious about expressing their opinions at the start of the debate and as the debate goes by they get more spontaneous and less conscious.

\section{Proposed Benchmark Model}
\begin{figure}[t]
    \centering
    \includegraphics[width=12cm]{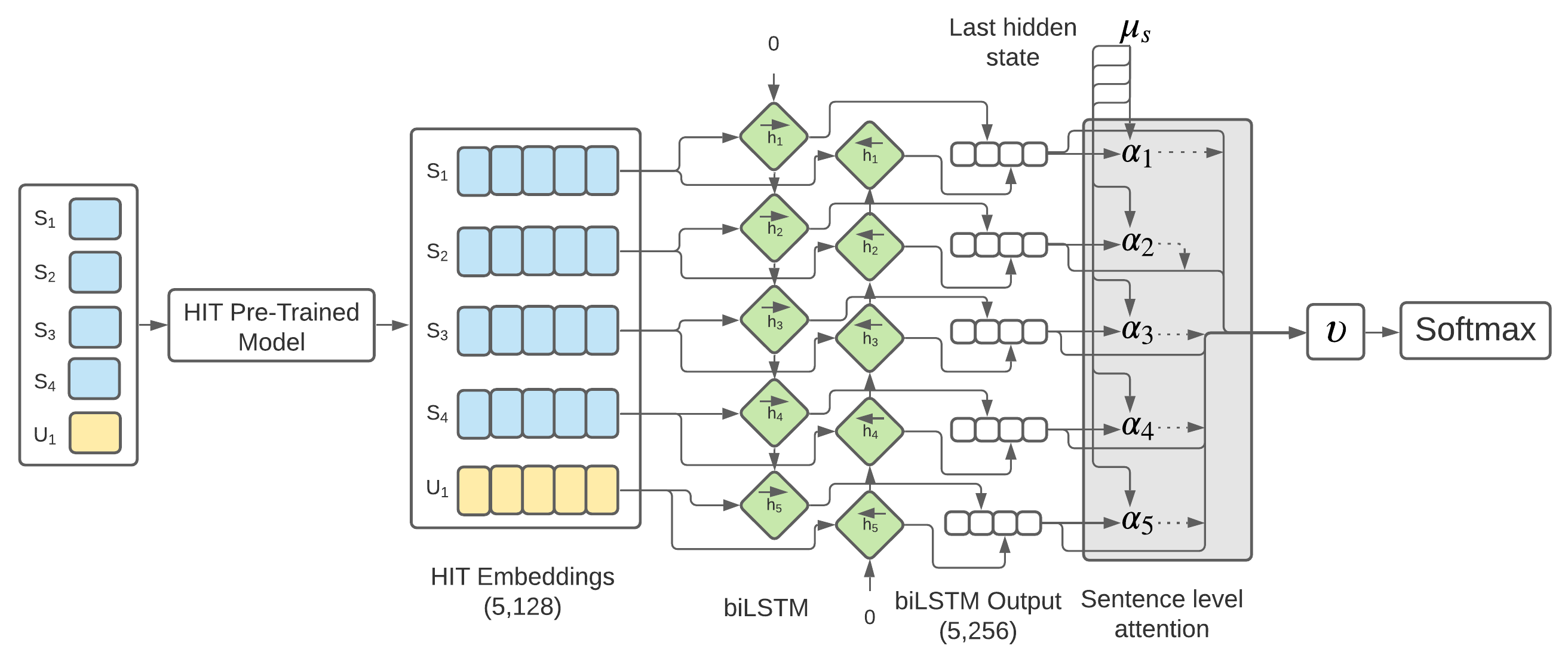}
    \caption{The proposed \model\ architecture for the anchor opinion detection.}
    \label{fig:model}
    \vspace{-6mm}
\end{figure}

In this section, we describe our proposed benchmark model, \model\ that we adopt for the anchor opinion detection task. Since the number of the opinionated anchor utterances are significantly few compared to the total number of utterances in the dataset, we adopt an instance-based modeling for the detection. For each anchor's utterance $u_t$, we create an instance that contains all previous utterances ($u_1, u_2, \cdots, u_{t-1}$) of the dialog as context and the target utterance $u_t$ as the last utterance of an instance. For each instance, we aim to classify the target utterance as opinionated or non-opinionated. We hypothesize that the fixed context will provide appropriate clue about the debate and, at the same time, restrict the model not to overwhelm itself in comprehending the desired context rather than focusing on the opinion discovery. A high-level architecture diagram for the anchor's opinion detection task is depicted in Figure~\ref{fig:model}.     

We feed each instance one-by-one to \model\ as input. Since code-mixed texts are susceptible to the spelling variations and various other kinds of noise, representations learned at the sub-word level often counter such variation quite efficiently. Recent literature shows that a wide range of character and sub-word level code-mixed representation models outperform word-level representation models for numerous tasks. Some of them are HIT \cite{sengupta-etal-2021-hit}, CS-ELMO \cite{aguilar-solorio-2020-english}, CNN\_LSTM \cite{joshi2016towards}, etc. We employ HIT (Hierarchically attentive Transformer), the most recent and robust representation learning method for code-mixed text among them, to capture the semantic and syntactical features of the debate. It encodes the code-mixed utterance in the embedding space where the semantic difference among various spelling variations of the same word is minimal. 

We obtain representation for each utterance in an instance and feed them through a biLSTM layer for sequence learning. The biLSTM layer captures the cross-sentence relationships across the utterances by exploiting the conversation dynamics of the dialog and subsequently learns latent representations $\overrightarrow{h_i}$ for each utterance $u_i$.
Next, we apply the multi-headed self-attention mechanism \cite{vaswani2017attention} to identify the importance of contextual utterances considering the target utterance $u_t$.
To this end, we treat the target utterance $h_t$ as the instance-level context vector $\mu_s$ and compute the interactions between the context vector and every utterance in the dialog through an interactive attention mechanism. The intuition is to obtain an abstract view of the instance that should help in exploiting the dialog dynamics corresponding to the target utterance in a better way. Subsequently, we accumulate the attention weights through a weighted summation and obtain the final vector as $v$.
\begin{eqnarray}
\hat{h_i} & = & tanh(h_i) ; \quad
\alpha_i = \frac{\exp{(\hat{h_i}^T\mu_s)}}{\sum_j \exp{(\hat{h_j}^T\mu_s)}} ; \quad
v = \sum_i \alpha_i \hat{h_i} \nonumber
\end{eqnarray}

Finally, we feed the vector $v$ to the softmax classifier for classifying the target utterance as opinionated or non-opinionated.

\section{Experiments and Results}
In this section, we report our experimental results and error analysis.  
\paragraph{\bf Baselines.}
Since opinion mining in code-mixed conversations is relatively unexplored arena, we include various code-mixed representation learning-based system as our baselines. In particular, we employ multi-lingual BERT (mBERT) \cite{bert}, XLM-RoBERTa \cite{xlm}, and IndicBERT \cite{kunchukuttan_ai4bharat-indicnlp_2020} based embedding models to extract the utterance representation. Subsequently, we fine-tune each of these systems through a biLSTM layer followed by a linear layer with softmax classification.

\paragraph{\bf Experiment Setup.}
Since \dataset\ is skewed towards the non-opinionated anchor's utterance, we perform oversampling at the instance-level for the opinionated utterances and obtain the equal number of opinionated and non-opinionated instance. For experiments, we perform 3-fold cross-validation and report the average for each case. Note that the oversampling is performed only for the training set in each fold. All experiments are performed on a 12GB K80 Tesla GPU server. 

For creating an instance, we vary the size of context from 1 to 7 and observe the best performance with context 5, i.e., $\langle (u_{t-5}, u_{t-4}, u_{t-3}, u_{t-2}, u_{t-1}), u_{t}\rangle$ as an instance. Furthermore, during experiments, we face a subtle challenge in obtaining the utterance representation for lengthy utterance (number of tokens $>512$), since most of the pre-trained language models (PLM) do not comprehend sentences more than $512$ tokens. In such cases, a typical solution is to clip the utterance at index $512$. However, in this work, we exploit an alternative without omitting the content. We split the lengthier utterances into $k$ chunks of $512$ tokens. Subsequently, we obtain representations for each of these $k$ chunks and consolidate them by taking an average of the $k$ representations. 

\paragraph{\bf Results and Comparative Study.}
We report the results of \model\ along with other baselines in Table \ref{tab:instance}. For each case, we compute the weighted-F1 scores. Moreover, we report the class-wise precision, recall, and F1 for the opinionated and non-opinionated cases as well. We observe that \model\ records the best weighted F1-score of 0.703 in comparison with 0.686 weighted F1-score of the best baseline, ML-BERT. Among all baselines, Indic-BERT has the least score at 0.657 weighted F1-score.

\begin{table*}[t]
\centering
\renewcommand{\arraystretch}{1.24}
\caption{Experimental results for anchor opinion detection.}
\resizebox{0.75\textwidth}{!}
{%
\begin{tabular}{l | c|c|c | c|c|c | c|c|c}
\multirow{2}{*}{\textbf{Model}}               & \multicolumn{3}{c|}{\textbf{Opinion}} & \multicolumn{3}{c|}{\textbf{Non-opinion}}   & \multicolumn{3}{c}{\textbf{Weighted}} \\ \cline{2-10}
\textbf{} &
  \textbf{F1} &  \textbf{Rec} &   \textbf{Pre} &  \textbf{F1} &  \textbf{Rec} &  \textbf{Pre} &  \textbf{F1} &  \textbf{Rec} &  \textbf{Pre} \\ \hline \hline
\textbf{ML-BERT} & 0.503 & 0.580 & 0.447 & 0.756 & 0.712 & 0.806 & 0.686 & 0.675 & 0.707 \\ \hdashline
\textbf{Indic-BERT} & 0.500 & \bf 0.630 & 0.458 & 0.723 & 0.669 & \bf 0.819 & 0.657 & 0.647 & \bf 0.724 \\ \hdashline
\textbf{XLM} & 0.424 & 0.468 & 0.468 & 0.759 & \bf 0.758 & 0.787 & 0.669 & 0.675 & 0.700 \\ \hline
\textbf{\model} & \bf 0.510 & 0.555 & \bf 0.471 &  \bf 0.778 &  0.752 &  0.806 & \textbf{0.703} &  \bf 0.692 &  0.715 \\ \hline
\end{tabular}%
}
\label{tab:instance}
\vspace{-4mm}
\end{table*}

We further observe the class-wise performance of all systems. For the opinion class, \model\ yields 0.510 F1-score, whereas, it obtains F1-score of 0.778 for the non-opinionated class. Similar to the weighted case, we obtain inferior results for all baselines in both classes. Another observation is that the performance for the non-opinionated class, irrespective of the model, is better than the opinionated class. We relate this to the complex nature of the anchor opinion detection task, where it is extremely challenging to comprehend the intended opinion especially in a conversational setup. 
\begin{table*}[t]
\centering
\caption{Token-level confusion matrix. We show performance w.r.t. a few critical words in anchors' utterances.}
\resizebox{0.8\textwidth}{!}{
\begin{tabular}{|P{0.69\textwidth}|P{0.065\textwidth}|P{0.065\textwidth}|P{0.065\textwidth}|P{0.065\textwidth}|}
\hline
\textbf{Words} & \textbf{TP} & \textbf{FN} & \textbf{TN} & \textbf{FP} \\ \hline
congress & 33 & 19 & 85 & 34 \\ \hline
bjp & 18 & 11 & 46 & 15 \\ \hline
modi & 24 & 16 & 77 & 27 \\ \hline
gandhi & 33 & 18 & 63 & 37 \\ \hline
bilkul & 14 & 15 & 45 & 20 \\ \hline
hindu & 14 & 10 & 50 & 20 \\ \hline
muslim & 6 & 4 & 14 & 6 \\ \hline
Question-based (kyu, kya, kab, kaha, kaun, kitne, kaise) & 70 & 47 & 235 & 102 \\ \hline
\end{tabular}}
\label{tab:words_preds}
\vspace{-3mm}
\end{table*}

\paragraph{\bf Error Analysis.}

In this section, we both quantitatively and qualitatively analyse of the results obtained from \model. 
As we observe in Table~\ref{tab:instance}, a relatively lower F1-score for the opinionated class suggests a significant number of false positives and false negatives. Moreover, we also observe a relatively higher false positives than the false negatives, thus reporting an inferior precision score. This could be due to presence of a few words which are highly inclined towards one class of utterance, as depicted in Figure \ref{fig:top10OP}.

Therefore, in Table~\ref{tab:words_preds}, we  investigate the words that were prevalent in the dataset, and their distribution in the results we obtained. We observe that utterances with reference to the two major Indian political parties (\textit{viz.} `congress' and `bjp') caused more false positives than false negatives.
On the other hand, question-based utterances (ones that usually start with words like `kyu' (why), `kya' (what), `kab' (when), `kaha' (where), `kitne' (how many), `kaise' (how)) have a very high false negative rate as compared to the false positive. For other words like `modi' and `gandhi' (who are political figures) proportionally have similar false positive and negative values. We also observe similar trends for the words representing two major religions in India, e.g., `hindu', `muslim', etc. 

In Table~\ref{tab:dataset:example:fn}, we report two mis-classified instances -- one for the false positive, while another for the false negative. We speculate that, in the first case, \model\ focuses too much on the contextual utterances and due to the presence of allegations in the second utterance of the instance, it classifies the instance as opinionated rather than non-opinionated. On the other hand, in the second example, a small portion of the target utterance (i.e., `tabhi yeh haal hai') suggests an opinion, which the model could not comprehend as the opinionated instance. Moreover, we observe a few mis-classified examples that are at the opposite end of the spectrum -- one requires context to get a sense of opinion whereas, for others, context is creating noise in the model. Such observation also signifies the subtleness of the proposed task.

\begin{table*}[t]
    \centering
    \caption{Examples of test instances which were wrongly predicted by \model.}
    \label{tab:dataset:example:fn}
    \renewcommand{\arraystretch}{1.24}
    \resizebox{\textwidth}{!}{%
    \begin{tabular}{cc|c|p{38em}|c|c}
        
        \multirow{35}{*}{\rotatebox{90}{Instance\#}} & & \multicolumn{2}{|c|}{Debate} & Gold & Predicted \\ \hline \hline
        & \multirow{10}{*}{1 } & \multirow{4}[20]{*}{Context} & [\textbf{A}] \textit{``Aapko kya lagta hai $<$name1$>$ nahi toh $<$name2$>$?''} (What do you think, if it is not sonia then it is rahul?) & & \\ 
        & & & [$\mathbf{S_5}$] \textit{``jeet haar ek vishe hai aaj jeet ke bhi log haar ja rahe hai kyu aap main wahi bata raha hun aapki baat ka jawaab de raha hun jeet haar ek vishe hai aaj log jeet ke bhi haar jaate hai bjp mla kharid leti hai''} (Winning and losing is one topic, today, even after winning people are losing. I am answering to your question only. Winning and losing is one topic, today, even after winning people are losing, bjp buys mlas (Member of the Legislative Assembly).) & & \\ 
        & & & \textbf{[A]} \textit{``nahi nahi aapko kya lagta hai $<$name1$>$ ya $<$name2$>$ aapko kya lagta hai $<$name1$>$ $<$name1$>$ ke baad ab $<$name2$>$''} (No no what do you think? $<$name1$>$ or $<$name2$>$, what do you think? Will it be $<$name2$>$ after $<$name1$>$?) & & \\ 
        & & & [$\mathbf{S_5}$] \textit{``isliye toh loktantr ka loktantr ka ki hatya kar di hai lekin ab sawaal yeh paida hota hai ki woh chahe rahul ho chahe sonia ho koi varshit neta ya jo bhi log congress ke bhavishay mein us kursi ko sambhalenge woh vishe abhi baad mein aata hai''} (This is the reason for the death of democracy. The question is be it rahul or sonia or some other politician or anyone who will lead and take the responsiblity of that position in congress in the future, this topic is of future.)  & & \\ \cdashline{3-6}
        & & Target & \textbf{[A]} \textit{``$<$name1$>$ ke $<$name1$>$ ke baad ab laut ke party $<$name2$>$ pe aayegi''} (Will the party fallback to $<$name2$>$ after $<$name1$>$ leaves?) & No & Yes \\ \cline{2-6}

        & \multirow{10}{*}{2 } & \multirow{4}[20]{*}{Context} & [\textbf{A}] \textit{``$<$name$>$ thik thik $<$name$>$ $<$name$>$ $<$name$>$''} ($<$name$>$ fine. fine. $<$name$>$, $<$name$>$, $<$name$>$) & & \\ 
        & & & [$\mathbf{S_1}$] \textit{``Ji''} (Yes.) & &  \\ 
        & & & \textbf{[A]} \textit{``Ji Ji toh main supreme court ko argue kar raha hun aapko supreme court pe bharosa nahi hai aapko air chief marshal par bharosa nahi hai thik hai $<$name$>$ thik''} (Yes, Yes, I am arguing the supreme court. Don’t you trust the supreme court? Don’t you trust the air chief marshal? Ok $<$name$>$) & &  \\ 
        & & & [$\mathbf{S_6}$] \textit{``apna jo opposition hai woh weak hai the congress was weak''} (Our opposition is weak. Congress was weak.)  & & \\ \cdashline{3-6}
        & & Target & \textbf{[A]} \textit{``\textcolor{blue}{sir yeh weak hai tabhi yeh haal hai} chaliye thik hai $<$name$>$ main aapke paas aa raha hun $<$name$>$ kya kya rafal ko thik kya rafal ko bofors banane ki koshish ho rahi hai kya yeh rafal ko bofors banane ki koshish hai kya $<$name$>$''} (Sir, it is weak that’s why the conditions are like this. Anyway, $<$name$>$ I am coming over to you. Is this an attempt to make Rafael deal like Bofors scam?) & Yes & No\\ \hline
        
    \end{tabular}%
    }
    \vspace{-3mm}
\end{table*}

\section{Conclusion and Future Work}
In this work, we proposed a novel task of anchor's opinion detection in code-mixed conversations. To this end, we curated \dataset, a first of its kind dataset by transcribing various debate videos from mainstream Indian news channels. We performed extensive analyses on \dataset, and reported interesting findings. Furthermore, we benchmark the \dataset\ dataset using \model\ -- an interactive attention-based framework on top to several pretrained code-mixed representation models. Moreover, we conducted error analysis on the outputs of \model.
In future work, we plan to extend the dataset with more opinionated samples as well as other varieties of debates. We also wish to explore the multi-modality for the opinion detection.

\section*{Acknowledgement}
The authors would like to acknowledge the support of the Ramanujan Fellowship (SERB, India), Infosys Centre for AI (CAI) at IIIT-Delhi, and ihub-Anubhuti-iiitd Foundation set up under the NM-ICPS scheme of the Department of Science and Technology, India.

\bibliographystyle{splncs04}
\bibliography{anthology}

\end{document}